\documentclass{bmvc2k}
\usepackage{dsfont}
\pdfoutput=1

\title{\textit{Look and Modify}: Modification Networks for Image Captioning}

\addauthor{Fawaz Sammani}{fawaz.sammani@aol.com}{1}
\addauthor{Mahmoud Elsayed}{elsayedmahmoud@aol.com}{1}

\addinstitution{
Multimedia University\\
Jalan Ayer Keroh Lama\\
75450 Melaka\\
Malaysia
}

\runninghead{FAWAZ SAMMANI, MAHMOUD ELSAYED}{LOOK AND MODIFY}

\begin{document}

\maketitle

\begin{abstract}
Attention-based neural encoder-decoder frameworks have been widely used for image captioning. Many of these frameworks deploy their full focus on generating the caption from scratch by relying solely on the image features or the object detection regional features. In this paper, we introduce a novel framework that learns to modify existing captions from a given framework by modeling the residual information, where at each timestep the model learns what to keep, remove or add to the existing caption allowing the model to fully focus on "what to modify" rather than on "what to predict". We evaluate our method on the COCO dataset, trained on top of several image captioning frameworks and show that our model successfully modifies captions yielding better ones with better evaluation scores. Code is at  \textcolor{red}{\textit{https://github.com/fawazsammani/look-and-modify}}

\end{abstract}

\section{Introduction}
\label{sec:intro}
Image captioning is the action of briefly describing an image in natural language, which lies at the junction of computer vision and natural language processing. It can be applied to many real-word applications such as human-machine interaction \cite{Das2017VisualD}, content-based image retrieval and assisting the visually impaired people. With the rise of deep learning, neural-based encoder decoder frameworks \cite{Vinyals2015ShowAT}  have proven to be highly effective, achieving significant results compared to previous image processing-based techniques.

After the introduction of attention mechanisms \cite{Bahdanau2015NeuralMT}, visual attention-based methods \cite{Xu2015ShowAA}\cite{Yang2016StackedAN} \cite{Yang2016EncodeRA} have been widely adopted in image captaining, where the model focuses on specific image regions when generating each word in the caption. Modern image captioning frameworks also incorporate object detection techniques into image captioning \cite{Lu2018NeuralBT} \cite{Anderson2018BottomUpAT}, where attention is also computed over the regions detected by an object detection framework, which visually grounds the words to their associated pixels in the image.  

However, all of these methods rely solely on the image and tend to refer to specific pixels (whether spatial maps or objects) when constructing each word, resulting in a scratch generation of the overall caption which is fully-dependant on the image and what has been generated from the caption so far. Moreover, these methods don't use any previous knowledge on what mistakes have been performed earlier, and what challenges the model has faced during prediction. In other words, they tend to fully focus on "what to predict".

To mitigate all of the problems mentioned, we introduce a novel framework that learns to correct captions from a previously-trained model by fully focusing on modeling "what needs to be changed" to the existing caption. This is achieved by modelling the residual information that needs to be added to the existing caption. Our model can be thought of as a modification network from what the decoder already knows about the existing caption. At each timestep, the model predicts the residual information which is the output of a language model, as well as a modification gate that represents how much information to take from the existing caption. Figure \ref{demo} demonstrates our modification model. To visualize our modification gate, we take the average of all output values in Equation 12 discussed in Section 3.5. Lower values of the modification gate correspond to more modification needed to the existing caption. Consider the words "cake" and "table" (left) and the words "standing" and "snow" (right). These words have a low modification gate value, since they were never seen in the existing caption, neither are semantically similar to any word in the existing caption. Moreover, the word "bear" (right) is never seen in the existing caption, however is semantically similar to "cats" in the existing caption, implying that "bear" can be inferred from the semantic meaning of the existing word which corresponds to an animal. 

To best of our knowledge, this is the first work that proposes a modification approach to image captioning. In summary, the overall contribution of this paper are presented as follows: 

\begin{figure}
    \centering
    \includegraphics[width=0.95\textwidth]{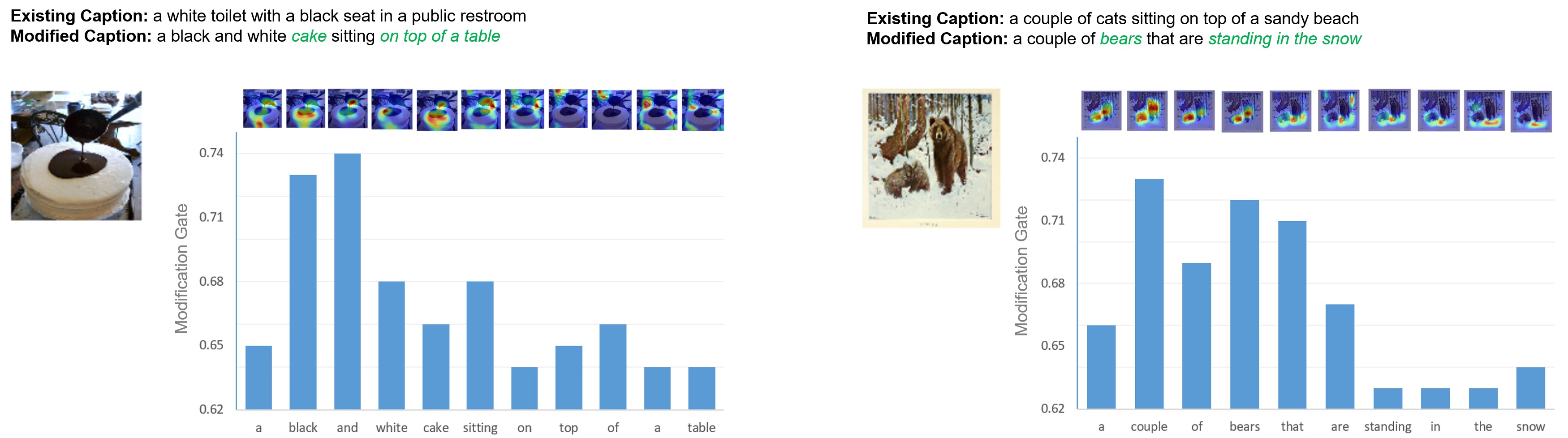}
    \caption{Our model learns to modify captions of an existing framework by using a modification gate. Lower values correspond to more modification needed to the existing caption. For example, the words "cake", "table", "standing" and "snow" were never seen in the existing caption, neither are semantically similar to any word in the existing caption.}
    \label{demo}
\end{figure}

\begin{itemize}
\item We transfer knowledge of captions from a previously trained model where at each timestep, the model keeps or modifies the word from the existing caption by using a modification gate. This eliminates the need of constructing the caption from scratch, and allows the model to fully focus on "what to modify" rather than on "what to predict". If the existing caption is mostly correct, but needs a one-word modification, then the modification networks corrects that word and leaves the remaining of the caption unchanged. Moreover, the modeled sentence embedding contains semantic information and is able to capture non-visual features and contribute to yield more accurate captions.

\item Our model can be plugged in on top of any image  captioning framework and trained separately or jointly to produce better results. We show that out model successfully modifies captions of several frameworks to better ones. This also gives the advantage of boosting the results of any later frameworks to be proposed. 
 \end{itemize}

\section{Related Work}
In general, image captioning models are divided into three categories, including neural-based methods \cite{Vinyals2015ShowAT}\cite{Mao2015DeepCW}\cite{Karpathy2015DeepVA}, attention-based methods \cite{Xu2015ShowAA}\cite{Yang2016StackedAN}\cite{Yang2016EncodeRA}\cite{You2016ImageCW} and RL-based methods \cite{Rennie2017SelfCriticalST}\cite{Ren2017DeepRL}. Recently, attention-based methods have been widely adopted, and have proven to be more effective than neural-based methods. Attention-based methods tend to focus on specific regions in the image when predicting each word in the caption. The first attention-based image captioning framework was proposed in \cite{Xu2015ShowAA}, where a weighted encoded image which includes visual information on parts of the image for a particular word, is generated at each timestep and guides the decoding process. This framework also has two variants, namely Soft-Attention and Hard-Attention. Following this work, many research on attention-based image captioning has been performed, some that proposed a semantic attention mechanism \cite{You2016ImageCW}, where top-down visual features are first extracted from the image and visual concepts including attributes and objects are then detected. Both the visual concepts and visual features are combined to produce a caption. This is not the only work on using attributes in image captioning. \cite{Yao2017BoostingIC} shows that image captioning may be boosted with attributes when supplied as an additional input to the LSTM. Other works have proposed an adaptive attention mechanism \cite{Lu2017KnowingWT} where attention is eliminated over non-visual words that can be computed from the language model itself, making the model attend to the "visual sentinel" rather than attending to the image regions.  

Most recently, top-down and bottom-up attention mechanisms \cite{Anderson2018BottomUpAT}  were introduced, where bottom-up features based on object detection regional outputs are obtained, and attention is calculated at the level of objects and salient image regions. Other works include template-based methods \cite{Lu2018NeuralBT}, which propose to visually-ground the generated word to outputs of object detection frameworks, which results in a natural language description explicitly grounded in regions found by object detectors.   

\section{Method}
We first describe the general attention-based neural encoder-decoder framework for image captioning in section 3.1, and then elaborate on our proposed methodology in sections 3.2, 3.3, 3.4, and 3.5.

\subsection{Attention-Based Neural Encoder-Decoder Frameworks}
Given an image \textbf{\textit{I}} with its corresponding caption \textbf{\textit{y}} which is represented as a series of words $\boldsymbol{\left\{y_{1}, y_{2},\ldots, y_{t-1}\right\}}$, using a recurrent-neural network (RNN) as the decoder, the neural-based encoder-decoder model maximizes the log likelihood of the RNN joint probability of each timestep \textbf{\textit{t}} which is obtained using the chain rule:

\begin{equation}
p(\boldsymbol{y} | \boldsymbol{I})=\prod_{t=1}^{T} p\left(y_{t} | \boldsymbol{y}_{1 : t-1}, \boldsymbol{I}, \boldsymbol{\theta}\right)
\end{equation}

With the superior performance of Long-Short term memory networks (LSTMs) and their strong capability of capturing long-term dependencies, the decoder RNN can be represented as an LSTM, where the hidden state at each timestep \textbf{\textit{t}} is modeled as:

\begin{equation}
\boldsymbol{h}_{t}=\operatorname{LSTM}\left(\boldsymbol{x}_{t}, \boldsymbol{h}_{t-1}, \boldsymbol{m}_{t-1}\right)
\end{equation} where $\boldsymbol{x}_{t}$ is the input to the LSTM at timestep \textbf{\textit{t}}, $\boldsymbol{h}_{t-1}$ is the previous hidden state, and $\boldsymbol{m}_{t-1}$ is the previous memory state at timestep $t-1$.

With the proposal of attention mechanisms, the context vector plays a crucial role in sequence modelling frameworks \cite{Sutskever2014SequenceTS} and significantly improves performance. The context vector  can be thought of as a focus element that guides the network when generating the prediction. In image captioning frameworks, at each timestep \textbf{\textit{t}} the context vector provides  visual information on where to look in the image in order to predict a word, rather than solely relying on a single hidden state. Thus, the decoder would attend to particular regions in the image during the caption generation process via the context vector which is computed as a weighted sum of each pixel in the spatial image obtained by the encoder CNN. 

\subsection{Modification Networks for Image Captioning}
We first give an overview of how our modification model is structured, and then elaborate on the model details. 

In a high overview, our model consists of two parts. The first part acts as a feature extractor and the second as a language modeler. The language modeler is a combination of two LSTMs, namely Attention LSTM and Language LSTM. The output of the feature extractor is obtained from a previously-trained model and encoded into a fixed size representation using a Deep Averaging Network (DAN), which is detailed in the following section, to produce a sentence embedding that is then used by the language modeler in the caption generation process. This results in our model being fully-differentiable, and can be trained in an end-to-end manner. A complete overview of our model is shown in Figure \ref{model}.

\subsection{Deep Averaging Network (DAN)}
    Given the output caption from the MLP layer of the existing model which is a sequence of word vectors $\boldsymbol{\left\{w_{t},\ldots, w_{m}\right\}}$, where \textbf{\textit{m}} represents a single caption length, our goal is to first construct a fixed-size feature vector for each caption of variable-length \textbf{\textit{m}}. To implement this, we make use of a Deep Averaging Network (DAN) \cite{Iyyer2015DeepUC}, which takes as input a variable-length sequence of data (in our case a sequence of word vectors), averages them all together and passes them through several linear layers with activation functions to produce a fixed-size representation of the input sentence at the last layer. The DAN can be mathematically described as follows:
\begin{equation}
a=\frac {1}{m}\sum_{i=1}^{m}{w_{i}}
\end{equation}
\begin{equation}
e_{1}=\tanh\left(W_{1} a+b_{1}\right)
\end{equation}
\begin{equation}
e_{2}=\tanh\left(W_{2}e_{1}+b_{2}\right)
\end{equation} where \textbf{\textit{m}} is the sentence length, $W_{1}$ and $W_{2}$ are the learnable parameters, and $b_{1}, b_{2}$ are the bias terms. The output of Eq.(3), $\boldsymbol{a}\in \mathcal{R}^{d}$ is the average of all the word vectors $\boldsymbol{w_{1 : m}}$. The output of the DAN $\boldsymbol{e_{2}} \in R^{d}$ represents the fixed-size sentence embedding. In our framework, we use a pre-trained DAN provided by \cite{Cer2018UniversalSE}. In the following sections, we will denote the sentence embedding $e_{2}$ as $e$ and show how it can be used in our model.  

\subsection{Attention and Language LSTM}
We use two separate LSTMs for attention and language modeling, similar to \cite{Anderson2018BottomUpAT}. The input to the first LSTM consists of the DAN output $e$, the mean pooled bottom-up features $v_{gb} = \frac{1}{k} \sum_{i} \boldsymbol{v}_{i}$ and the previous word embedding $w_{t}$, such that $x_{t}^{1} = [e ; v_{gb} ; w_{t}]$ where $;$ indicates concatenation. These inputs provide the Attention LSTM context about the global image features as well as an overview of the general context of the caption to be generated, where the model observes what is already known about the caption and therefore can successfully model the residual of its input. The output of the attention LSTM ${h}_{t}^{1}$ is then used to compute an attention weighted feature vector of the image features. Inspired by \cite{Li2017ImageCW}, we add a global image feature to the local image features to allow attention over the global image information, resulting in image features $\boldsymbol{V}=\left\{v_{1}, v_{2}, \ldots, v_{p}, v_{g}\right\}$ where $v_{i} \in \mathcal{R}^{2048}$ and $p$ is the number of pixels. In particular, we use the output of the the last convolutional layer of a ResNet-101 with dimensions $2048 \times 8 \times 8$ and use an adaptive pooling mechanism to reshape the spatial output to have dimensions $2048 \times 14 \times 14$. The attention weights are computed by: 

\begin{figure}
    \centering
    \includegraphics[width=0.95\textwidth]{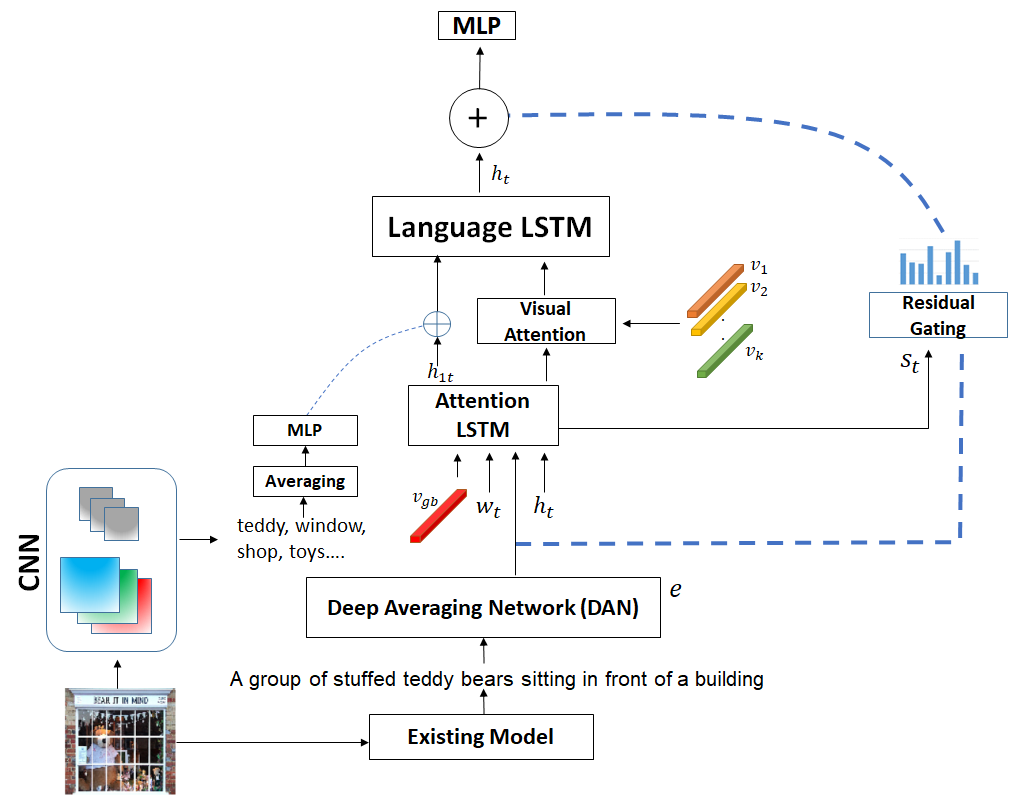}
    \caption{The proposed architecture of our modification network. The caption is first obtained from a previously trained framework and encoded using a Deep Averaging Network into a sentence embedding of fixed size representation. The residual information is then modeled by two LSTMs. The output of the language LSTM models the residual information to be added to part of the sentence embedding obtained by the residual gating block which outputs a modification gate.}
    \label{model}
\end{figure}

\begin{equation}
\alpha_{t_l}=\operatorname{softmax}(w_{s}^{T} \operatorname{Re} \operatorname{LU}\left(\boldsymbol{W}_{v} \boldsymbol{V}+\boldsymbol{W}_{hl} \boldsymbol{h}_{t}^{1}\right))
\end{equation}
The attention weighted image can then be computed by:
\begin{equation}
c_{t_{l}}=\sum_{i=1}^{p+1} \alpha_{i t_{l}} v_{i}
\end{equation}
In a similar manner, when using $k$ = 36 bottom-up features, we compute the attention weights over $\boldsymbol{B}=\left\{b_{1}, b_{2}, \ldots, b_{k}\right\}$ using:
\begin{equation}
a_{t_b}=\operatorname{softmax}(\boldsymbol{w}_{a}^{T} \operatorname{ReLU} \left(W_{vb} \boldsymbol{B}+W_{hb} \boldsymbol{h}_{t}^{1}\right))
\end{equation}

The attended bottom-up features are then computed by: 
\begin{equation}
c_{t_{b}}=\sum_{i=1}^{k} \alpha_{i t_{b}} b_{i}
\end{equation}
The attribute model is a CNN applied to image regions which is used to train visual detectors for words that frequently occur in captions using multiple-instance learning. This network outputs a set of words which can be nouns,verbs or adjectives. We follow the technique proposed in \cite{Fang2015FromCT} for  attribute  extraction. After all attributes have been extracted, we select the top-5  attributes  with  the  highest  scores.  The  5  attributes $A = \left\{a_{1}, a_{2}, \dots, a_{5}\right\}$ are averaged and passed through two linear layers with tanh activation function to reduce their dimensions, and serve as input to the Language LSTM such that $x_{t}^{2} = [A_{avg} ; h_{t}^{1} ; c_{t}]$ where $;$ indicates concatenation. The language LSTM thus receives information about the weighted image, the output of the attention LSTM and the attributes, providing it with sufficient information to construct the appropriate word. The output of the Language LSTM is given by $h_{t}$. Note that $c_{t}$ is either $c_{t_l}$ or $c_{t_b}$. In particular, we use the context vector $c_{t_l}$ when modifying captions from the soft-attention framework \cite{Xu2015ShowAA} and the adaptive attention framework \cite{Lu2017KnowingWT}, and $c_{t_b}$ when modifying captions from the bottom-up and top-down attention framework \cite{Anderson2018BottomUpAT}. For more details on the frameworks used, see section 6.1.
\subsection{Residual Gating}
As mentioned earlier, the DAN encodes the existing caption into a sentence embedding of a fixed size representation. Therefore, the encoded sentence represents all the caption words generated. However, the model is operating on timesteps (i.e. predicting one word at a time). Therefore, it is necessary to learn what to keep and what to remove from words of the existing caption at each timestep to result in an informative noun phrase or single word to be added to the residual information $h_{t}$ at that particular timestep. Therefore, we introduce a residual gating block that outputs a modification gate which is then multiplied with the sentence embedding to produce the respective part of the encoded sentence. This gate is similar to an LSTM gate, and can therefore be considered as an extension to an LSTM cell. In order to compute the latent state of what the decoder already knows about what has been generated so far, we make use of the visual sentinel proposed in \cite{Lu2017KnowingWT} which is an extension to the LSTM that separately models the information stored in the memory cell. The visual sentinel $s_{t}$ and its gate $g_{s}$ are modeled as:

\begin{equation}
g_{s}=\sigma\left(\boldsymbol{W}_{x} \boldsymbol{x}_{t}+\boldsymbol{W}_{h} \boldsymbol{h}_{t-1}\right)
\end{equation}
\begin{equation}
\boldsymbol{s}_{t}=\boldsymbol{g}_{s} \odot \tanh \left(\boldsymbol{m}_{t}\right)
\end{equation}

We then compute the modification gate by finding the similarity between what the decoder already knows (i.e. $s_{t}$) and the existing caption (i.e. $e$) through simple MLP layers, and passing the similarity output to a sigmoid activation function $\sigma$:
\begin{equation}
g_{r}=\sigma\left(W_{e} e + W_{m} \boldsymbol{s}_{t}\right)
\end{equation}

The output of the modification gate is a vector of dimension $d$ with each of its components represented as a scalar value between 0-1. By performing an element-wise multiplication of the gate with the sentence embedding, we obtain part of the existing caption $w_{f_{t}}$ which is added with the residual information $h_{t}$ to produce an output $o_{t}$ at each timestep:
\begin{equation}
w_{f_{t}}=g_{r} \odot e
\end{equation}
\begin{equation}
o_{t}=w_{f_{t}}+ h_{t}
\end{equation}
where we resize the hidden output $h_{t}$ to have dimension $d$ using a single layer neural network with tanh activation function.
The probability over the vocabulary of words is then computed by:
\begin{equation}
p_{c}=\operatorname{softmax}\left(W_{p} o_{t}\right)
\end{equation}

\section{Objective}
Our loss function is a combination of two terms. The first term is the general cross-entropy loss. The second term is denoted as the attribute loss. Rather than limiting the advantage of the attributes to only supplying them as an averaged input to the language LSTM, we leverage the attributes to act as a supervisor role during training time. We therefore maximize the objective of all extracted attributes being present in the caption. Given a generated caption of length \textbf{\textit{m}} represented as a sequence of words: $\boldsymbol{\left\{y_{1}, y_{2},\ldots, y_{m}\right\}}$ and a sequence of attributes extracted from the image \textbf{\textit{I}}: $\boldsymbol{\left\{a_{1}, a_{2},\ldots, a_{5}\right\}}$, we compute the occurrence \textbf{\textit{f}} of all attributes in the caption for a mini-batch. We then model the attribute loss term by taking a scaled negative exponential function of \textbf{\textit{f}} normalized across all samples. Notably, when the occurrence \textbf{\textit{f}} is zero (i.e. no attributes present in the generated caption), the term will be at its maximum. However, our observation is that the attribute loss term does not significantly improve performance. 
\begin{equation}
\begin{split}
&-\frac{1}{N} \sum_{i=1}^{N} \sum_{t=1}^{T} \log \left(p\left(y_{t} | y_{1 : t-1} ; I ; e\right)\right) +\beta\left(\mathds{1} e^{\frac{-2f}{3N} }\right)
\end{split}
\end{equation}
where $\mathds{1}$ is an indicator function which is true if the normalized occurrence $\frac{f}{N}<4$. $\beta$ is a coefficient that controls how much penalty of the attribute loss term to be considered. We set $\beta$ to be 0.4.

\section{Training Details}
In our setup, the attention and language LSTM are of single layer with a hidden size of 1000. We use the Adam optimizer \cite{Kingma2015AdamAM} with an initial learning rate of 5e-4 and anneal the learning rate by a factor of 0.8 every 3 epochs. We set the word embedding size to 1000, and the sentence embedding and attention size to 512. We use a batch size of 80 and start fine-tuning the encoder CNN for Soft-Attention and Adaptive Attention after 20 epochs with a learning rate of 1e-4. We do not fine-tune the encoder for bottom-up and top-down attention. We train for a maximum of 40 epochs with early stopping if the validation CIDEr score has not improved for 6 consecutive epochs. Our model can be trained in less than 48 hours on a single RTX2080Ti GPU. When sampling, we use a beam size of 3. We also use variational dropout to effectively regularize our language model, which samples one mask and uses it repeatedly across all timesteps. In that case, all timesteps of the language model receive the same dropout mask.

\section{Results and Analysis}

\begin{table}[ht]
\centering
\resizebox{\textwidth}{!}{\begin{tabular}{rlrrrrrrr}
Method                    & BLEU-1  & BLEU-2  & BLEU-3  & BLEU-4  & ROUGE-L    & CIDEr     & SPICE          \\ \hline
Soft-Attention \cite{Xu2015ShowAA}           & \multicolumn{1}{c}{70.1} & \multicolumn{1}{c}{49.2} & \multicolumn{1}{c}{34.4} & \multicolumn{1}{c}{24.3} & \multicolumn{1}{c}{-}    & \multicolumn{1}{c}{-}     & \multicolumn{1}{c}{-}    \\
Soft-Attention (ResNet) & \multicolumn{1}{c}{73.7} & \multicolumn{1}{c}{56.9} & \multicolumn{1}{c}{43.4} & \multicolumn{1}{c}{33.0} & \multicolumn{1}{c}{54.5} & \multicolumn{1}{c}{1.030}  & \multicolumn{1}{c}{19.1} \\ \hline
\textbf{Ours-MN (k=3)}                      & \multicolumn{1}{c}{\textbf{74.6}} & \multicolumn{1}{c}{\textbf{58.6}} & \multicolumn{1}{c}{\textbf{44.9}} & \multicolumn{1}{c}{\textbf{34.2}} & \multicolumn{1}{c}{\textbf{55.1}} & \multicolumn{1}{c}{\textbf{1.060}} & \multicolumn{1}{c}{\textbf{19.4}} \\ \hline
Method \\ \hline
Spatial \cite{Lu2017KnowingWT}    & \multicolumn{1}{c}{73.4} & \multicolumn{1}{c}{56.6} & \multicolumn{1}{c}{41.8} & \multicolumn{1}{c}{30.4} & \multicolumn{1}{c}{-}    & \multicolumn{1}{c}{1.024} & \multicolumn{1}{c}{-}    \\
Adaptive \cite{Lu2017KnowingWT}   & \multicolumn{1}{c}{74.2} & \multicolumn{1}{c}{58.0} & \multicolumn{1}{c}{43.9} & \multicolumn{1}{c}{33.2} & \multicolumn{1}{c}{54.9} & \multicolumn{1}{c}{1.052} & \multicolumn{1}{c}{19.4} \\ \hline
\textbf{Ours-MN (k=3)}    & \multicolumn{1}{c}{\textbf{75.1}} & \multicolumn{1}{c}{\textbf{59.1}} & \multicolumn{1}{c}{\textbf{45.3}} & \multicolumn{1}{c}{\textbf{34.5}} & \multicolumn{1}{c}{\textbf{55.4}} & \multicolumn{1}{c}{\textbf{1.071}} & \multicolumn{1}{c}{\textbf{19.6}} \\ \hline

Method                    \\ \hline
Top-Down (Cross Entropy) \cite{Anderson2018BottomUpAT} & \multicolumn{1}{c}{76.7} & \multicolumn{1}{c}{60.8}    & \multicolumn{1}{c}{46.8}    & \multicolumn{1}{c}{35.8} & \multicolumn{1}{c}{56.3} & \multicolumn{1}{c}{1.107} & \multicolumn{1}{c}{20.2} \\ \hline
\textbf{Ours-MN (k=3)}               & \multicolumn{1}{c}{\textbf{76.9}}  & \multicolumn{1}{c}{\textbf{61.2}}  & \multicolumn{1}{c}{\textbf{47.3}}  & \multicolumn{1}{c}{\textbf{36.1}} & \multicolumn{1}{c}{\textbf{56.4}} & \multicolumn{1}{c}{\textbf{1.123}} & \multicolumn{1}{c}{\textbf{20.3}} \\ \hline

\end{tabular}}
 \caption{Performance on the MSCOCO Karpathy test split reported on 7 evaluation metrices including BLEU-n where n is the number of grams, ROUGE-L, CIDEr and SPICE. Results are shown when our model is trained on top of 3 different image captioning frameworks and evaluated at beam size k = 3. Note that we do not use CiDEr optimization. Scores are higher in all evaluation metrics.} 
\end{table} 

\begin{figure}
    \centering
    \includegraphics[width=0.9\textwidth]{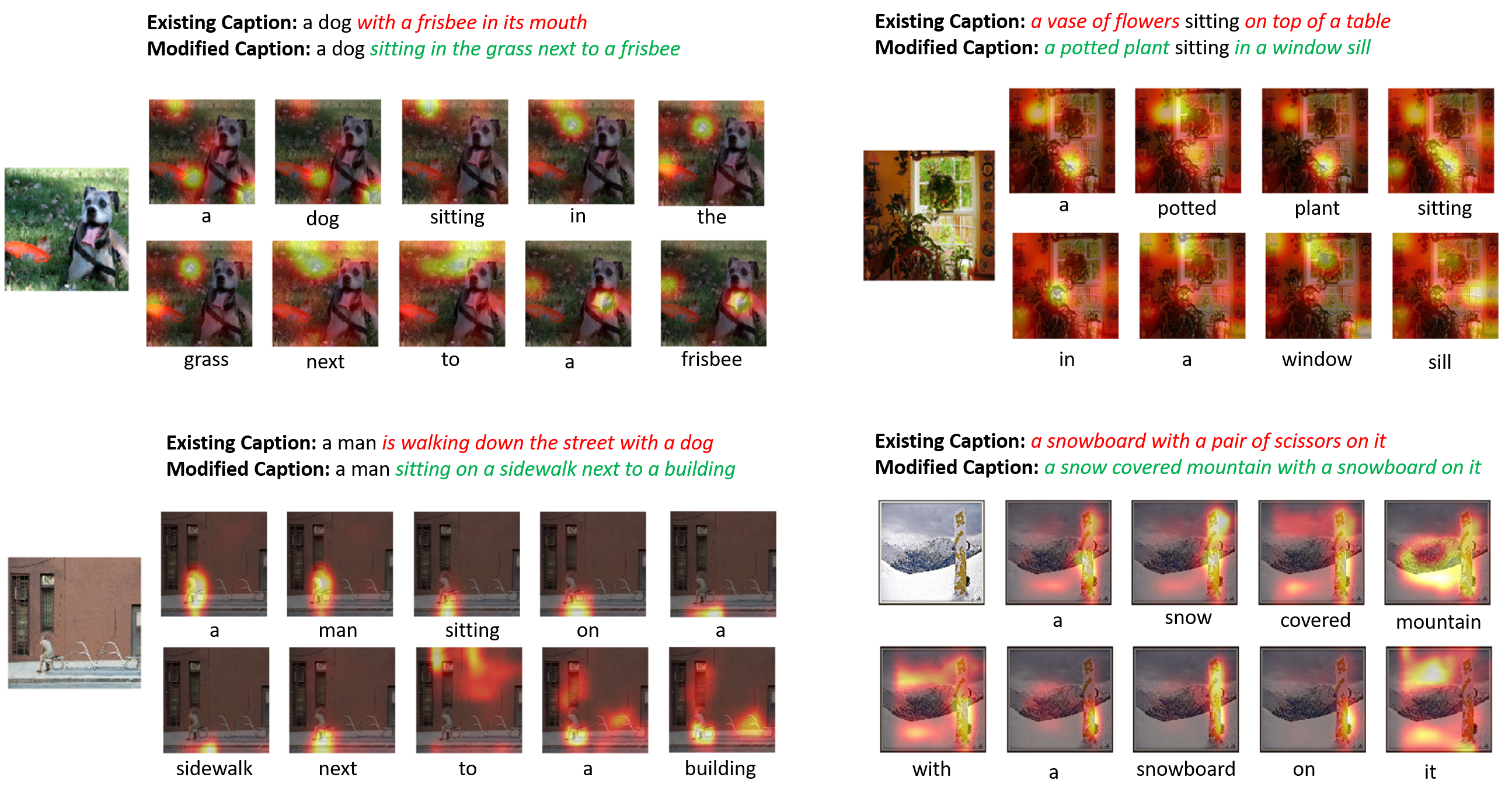}
    \caption{Results of complete modification resulting from our model. Most (or all) words of the existing caption are modified.}
    \label{complete}
\end{figure}

\begin{figure}
    \centering
    \includegraphics[width=0.9\textwidth]{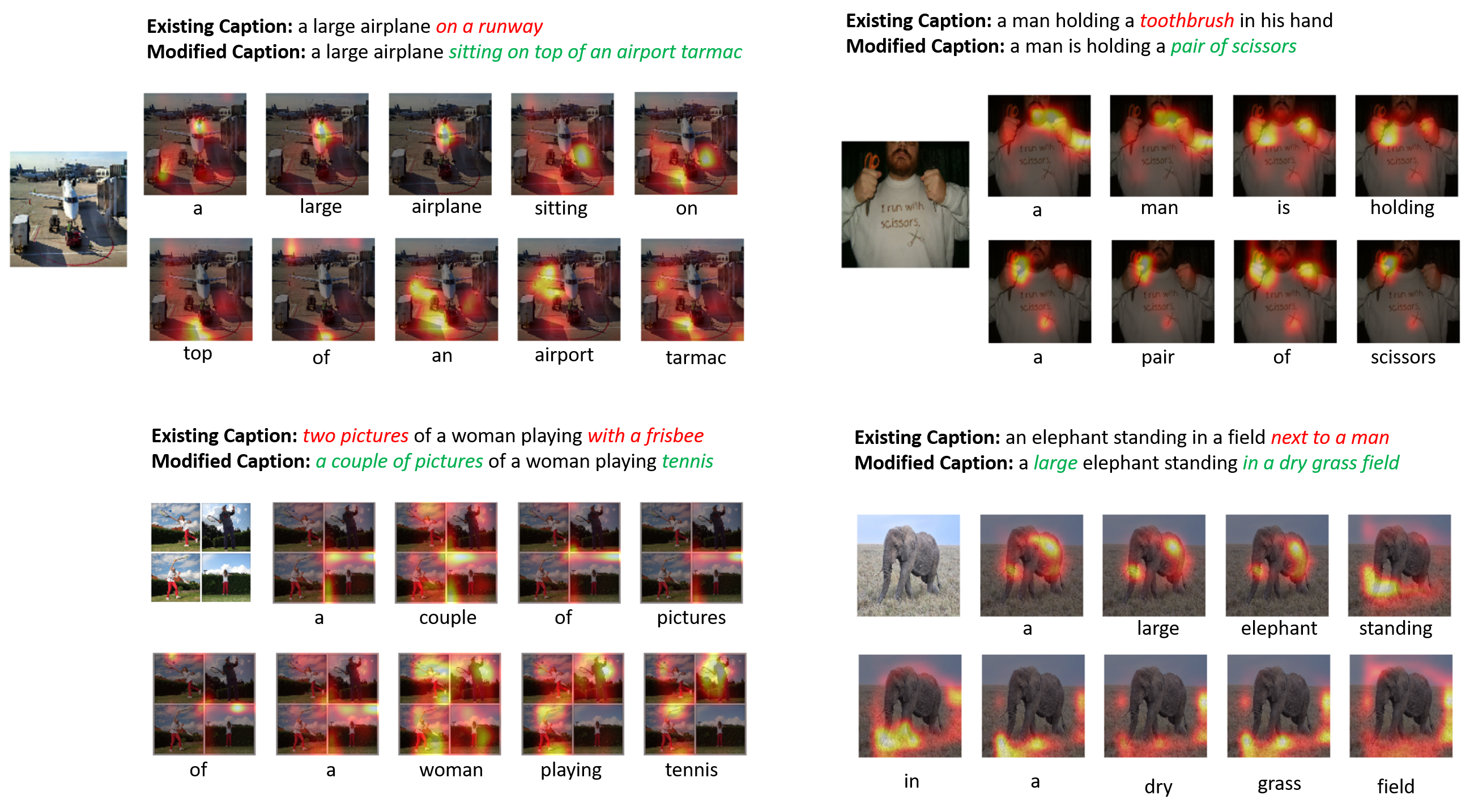}
    \caption{Results of partial modification resulting from our model. Parts or phrases from the existing caption are modified.}
    \label{partial}
\end{figure}

\begin{figure}
    \centering
    \includegraphics[width=0.9\textwidth]{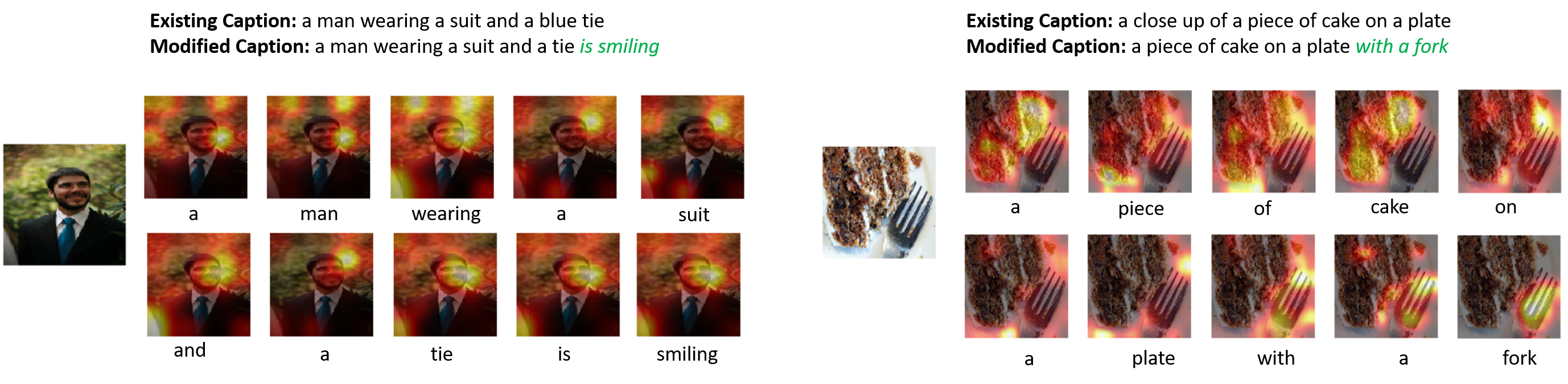}
    \caption{Results of complementary modification resulting from our model. The existing caption is completely correct, but lacks details and extra information to produce a more accurate description.}
    \label{comple}
\end{figure}

\subsection{Experimental Setup}
We run experiments on the Microsoft COCO dataset \cite{Lin2014MicrosoftCC}, which contains 82,783 training images, 40,504 and 40,775 images for validation and testing, respectively. Each image in the dataset contains 5 captions annotated by 5 different people. For fair comparison with other works, we use the data split in \cite{Karpathy2015DeepVA}, which contains 113,287 training images, and 5,000 images for validation and testing, each. We use a maximum caption length of 30, and build our vocabulary of words that appear at least 5 times in the training set, resulting in 9,490 words. 

We verify our modification network by training our model on 3 different image captioning frameworks, and evaluating on each. When training, the captions of the existing framework are generated from the training images and used as input to the DAN. Similarly, when evaluating, the validation and testing captions of the existing frameworks are taken to be the input to our DAN. The 3 frameworks used in this work are \textbf{Soft-Attention}\cite{Xu2015ShowAA}, \textbf{Adaptive Attention}\cite{Lu2017KnowingWT} and \textbf{Bottom-Up and Top-Down Attention}\cite{Anderson2018BottomUpAT}. During training, we don't fine-tune the previously-trained model weights. The output caption of the existing model is directly passed to the DAN.

\subsection{Quantitative and Qualitative Evaluation}
We utilize the COCO captioning evaluation toolkit \footnote{https://github.com/tylin/coco-caption}, and report experimental results on the following metrices: BLEU-n \cite{Papineni2001BleuAM} (BLEU-1 BLEU-2, BLEU-3, BLEU-4), Rouge-L \cite{Lin2004ROUGEAP}, CIDEr \cite{Vedantam2015CIDErCI} and the newly developed SPICE \cite{Anderson2016SPICESP} which is more close to human-level evaluation. Table 1 demonstrates our results on the MSCOCO Karpathy validation split when the model is trained on top of the 3 different image captioning frameworks. 
Scores of the original models are all reported after re-training these models using open-source implementations. We use ResNet-101 features for the Soft-Attention model \cite{Xu2015ShowAA} and the Adaptive Attention model \cite{Lu2017KnowingWT}, and bottom-up features for the bottom-up and top-down attention model.

To visualize our model, at each timestep of the caption generation process, we up-sample the attention weights to have the same size as the input image using nearest interpolation. We present three phenomenons of our modification networks, namely complete modification, partial modification and complementary modification. 

\textbf{Complete Modification} occurs when most (or all) words of the existing caption have been modified. An example of this phenomenon is shown in Figure \ref{complete}. The existing caption may succeed in understanding the general context of the image, but fails to arrange the objects, attributes and relations correctly. Another case is when the existing caption describes wrong objects and attributes which are not present in the image. Our model successfully modifies captions resulting from these two cases. 

\textbf{Partial Modification} occurs when a part or phrase from the existing caption is modified. In most cases, the existing caption is mostly correct, but includes a group of words or phrases that are false. An example of this phenomenon is shown in Figure \ref{partial}. 

\textbf{Complementary Modification} occurs when the existing caption is fully correct, but lacks some details and extra information to be present in order to produce a more accurate description of the image. An example of this phenomenon is shown in Figure \ref{comple}. 

\section{Conclusion}
We presented a novel image captioning modification framework that learns to modify any existing caption from a previously-trained model, focusing on "what to modify" from the caption rather than focusing on predicting the caption from scratch by relying solely on the image. Our model can be easily plugged in on top of any existing or future framework and trained separately or jointly to yield better results. We show that our model improves evaluation scores by experimenting with 3 different frameworks. Our model is not limited to image captioning, but can be applied on other tasks such as neural machine translation.

{\small
\bibstyle{ieee}
\bibliography{bmvc_final}
}

\end{document}